\documentclass[10pt,twocolumn,letterpaper]{article}

\usepackage{cvpr}
\usepackage{times}
\usepackage{epsfig}
\usepackage{graphicx}
\usepackage{amsmath}
\usepackage{amssymb}

\usepackage[breaklinks=true,bookmarks=false]{hyperref}

\cvprfinalcopy 

\def\cvprPaperID{****} 

\begin{document}

\title{Highlight Specular Reflection Separation based on Tensor Low-rank and Sparse Decomposition Using Polarimetric Cues}

\author{Moein Shakeri, Hong Zhang\\
Department of Computing Science\\
University of Alberta, Edmonton, AB, Canada\\
{\tt\small shakeri@ualberta.ca, hzhang@ualberta.ca}
}

\maketitle

\begin{abstract}
   This paper is concerned with specular reflection removal based on tensor low-rank decomposition framework with the help of polarization information. Our method is motivated by the observation that the specular highlight of an image is sparsely distributed while the remaining diffuse reflection can be well approximated by a linear combination of several distinct colors using a low-rank and sparse decomposition framework. Unlike current solutions, our tensor low-rank decomposition keeps the spatial structure of specular and diffuse information which enables us to recover the diffuse image under strong specular reflection or in saturated regions. We further define and impose a new polarization regularization term as constraint on color channels. This regularization boosts the performance of the method to recover an accurate diffuse image by handling the color distortion, a common problem of chromaticity-based methods, especially in case of strong specular reflection. Through comprehensive experiments on both synthetic and real polarization images, we demonstrate that our method is able to significantly improve the accuracy of highlight specular removal, and outperform the competitive methods to recover the diffuse image, especially in regions of strong specular reflection or in saturated areas.
\end{abstract}

\section{Introduction}

Objects in real world scenes have both diffuse and specular reflections, as a common physical phenomenon. On one hand, detecting the specular reflection can help us to infer the light direction, scene geometry~\cite{Related_12_2} and camera location. On the other hand, the existence of specular reflection presents difficulties for many applications including image segmentation, object detection~\cite{obj_detect, Intro6, Intro91}, pattern recognition~\cite{Intro9}, background subtraction~\cite{Intro4, Intro5, Intro7}, HDR reconstruction~\cite{Intro2, Intro3}, tracking~\cite{tracking, Intro8}, and 3D reconstruction~\cite{shakeri2021, Intro1}. Therefore, the specular reflection from images is a crucially important consideration in a variety of computer vision and robotics tasks. This line of research has been extensively studied, and many solutions proposed in the last decade on the benefit of chromaticity or polarization. However, almost all existing solutions have some significant drawbacks.

Methods based on the analysis of color space rely on either image statistics or strong prior assumptions~\cite{Related_4,Related_8} which are not robust in real scenes~\cite{Related_21}. Methods based on the theory of polarization use the fact that the specular reflection tends to be polarized while the diffuse reflection is unpolarized~\cite{Related_16,Related_18,Related_17}. However, in practice the specular reflection is partially polarized~\cite{Related_21}. Therefore, despite the promising result on some cases, these methods cannot obtain pleasing results while the assumption is violated.

\begin{figure}[t]
\centering
\includegraphics[width=\linewidth]{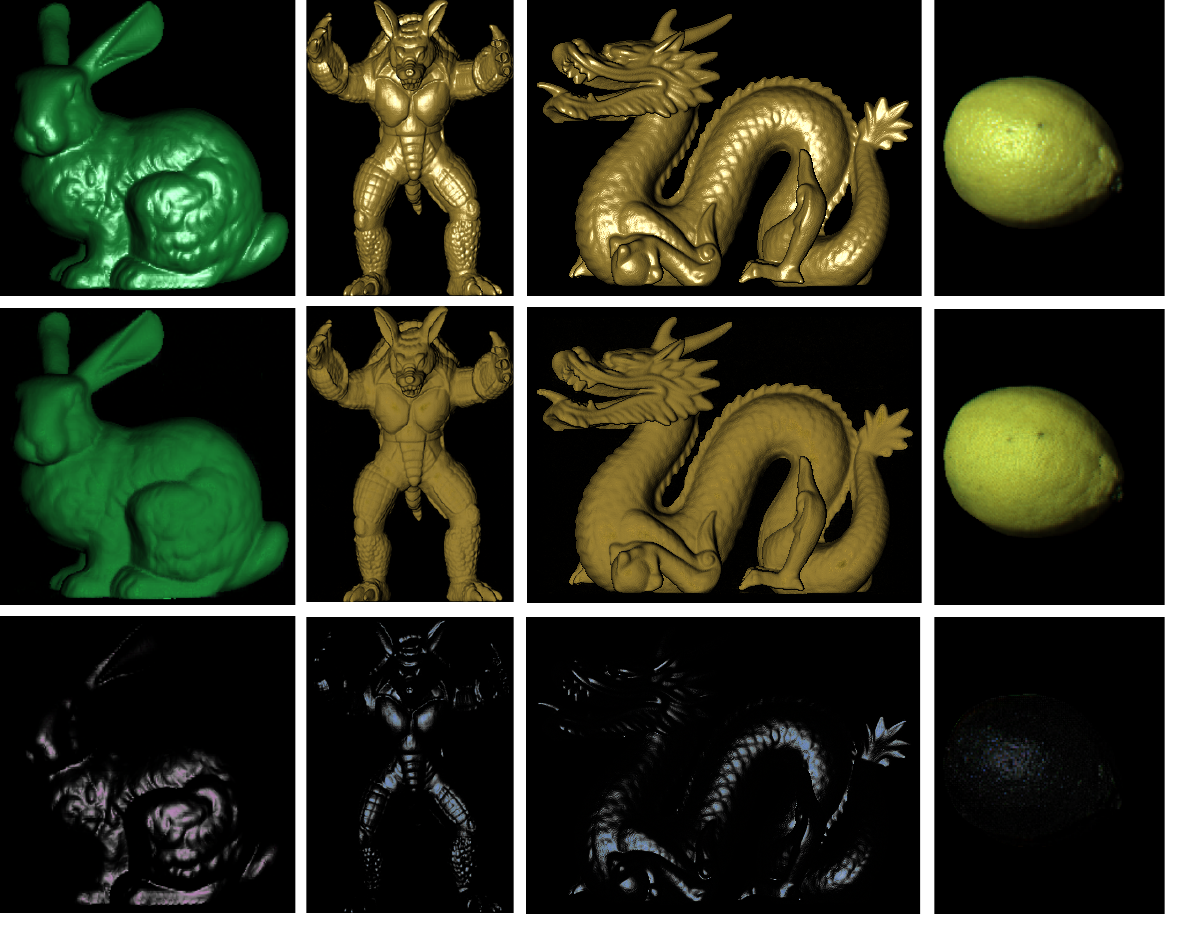}
\caption{Sample results of the proposed method. Rows from top to bottom: Input rgb image, recovered diffuse image, and sparse highlight specular image.}
\label{sample_res}
\end{figure}

In this paper, we propose an optimization based solution to separate highlight specular reflection from diffuse reflection using a tensor low-rank and sparse decomposition framework, with the help of polarimetric cues. Specifically, we first analyze the polarization images to compute the polarization chromaticity image. We then create a group of candidates for each pixel using the obtained polarization chromaticity image, and form all the candidates in a tensor structure as different representations of the original image. Based on polarization theory, we also define the phase angle constraint as a new polarization regularization term. We then introduce our proposed formulation by imposing it into the tensor low-rank and sparse decomposition framework. Finally, we optimize the proposed formulation to extract the diffuse images. 
We have evaluated the proposed method on both synthetic images and real images captured by a polarization camera. Sample results of our proposed method are shown in Fig.~\ref{sample_res}.

The main contributions of this paper are as follows.
\begin{itemize}
\item{We propose to use the polarization chromaticity image using polarimetric information.}
\item{We propose to create multiple representations for each pixel, and form them  as a tensor. In particular, we use block-processing to select some candidates for each pixel with the help of polarization chromaticity image, and then stack them in a 3D tensor data structure.}
\item{We introduce a phase angle regularization term using polarization images. Phase angle is only related to the surface geometry and so R, G, and B channels of each pixel should share a similar phase angle. We use this concept to form our phase angle regularization term.}
\item{We introduce an iterative optimization process by integrating the chromaticity-based specular removal and our phase angle regularization term into the tensor low-rank and sparse decomposition framework. To our knowledge, this is the first tensor-based approach using polarimetric cues to remove highlight specular reflections from images.}
\end{itemize}

The remainder of this paper is organized as follows. Related works on specular and diffuse reflection separation and the theory of polarization are summarized in Section~\ref{related_work}. Section~\ref{proposed_method} explains the details of our polarimetric highlight specular removal method. Experimental results and discussion are presented in Section~\ref{experiment}, and concluding remarks in Section~\ref{conclusion}.

\section{Related Works}
\label{related_work}

In this section we first review related works on specular reflection removal and then explain relevant polarization theory that is the basis of our proposed method.
 
\subsection{Highlight Specular Reflection Removal}

Specular reflection removal from images is a well-studied area of research and many solutions have been proposed which can be grouped into two major categories based on the number of images used. 

\vspace{-7pt}\subsubsection{\textbf{Single-Image Highlight Specular Separation Methods}}

Methods in this category remove specular reflection from a single image. Since this problem is inherently ill-posed, prior knowledge or assumption on the characteristics of natural images should be exploited to make the problem tractable. Most single-image-based methods require color segmentation~\cite{Related_1,Related_2} which is not robust for complex textured images or requires user assistance for highlight detection~\cite{Related_3}. To address this issue, Tan {\it et al.}~\cite{Related_4} proposed a method to remove highlights by iteratively shifting chromaticity values towards those of the neighboring pixels having the maximum chromaticity in the neighborhood without explicit color segmentation. Similarly, Mallick {\it et al.}~\cite{Related_5} proposed an SUV color space which separated the specular and diffuse components into S and UV channels. They used this SUV space to remove highlights by iteratively eroding the specular channel. This type of approaches may encounter problems due to discontinuities in surface colors, across which diffuse information cannot be accurately propagated. These methods also perform poorly on large specular regions.

Other approaches exist for single-image highlight specular removal and they pioneered by the idea of specular-free image~\cite{Related_4}, a pseudo-diffuse image that has the same geometrical profile as the true diffuse component~\cite{Related_6,Related_7,Related_8,Related_8_1,Related_8_2,Related_8_3,Related_8_4}. Kim {\it et al.}~\cite{Related_6} proposed an approximated specular-free image via applying the dark channel prior. Sue {\it et al.}~\cite{Related_7} defined $l_2$ chromaticity and used it to generate the specular-free image. Yang {\it et al.}~\cite{Related_8} proposed a fast bilateral filter adopting the specular-free image the range weighting function. The main drawback of the specular-free image is it suffers from hue-saturation ambiguity which exists in many natural images~\cite{Related_23}.

Inspired by the fact that highlight regions in many real-world scenes are contiguous pieces with relatively small size while colors of diffuse reflection can be well approximated by a small number of distinct colors, some methods~\cite{Related_22,Related_23} use low-rank and sparse decomposition framework. Akashi {\it et al.}~\cite{Related_22} formulate the separation of reflections as a sparse non-negative matrix factorization (NMF) problem. Since NMF is highly sensitive to outliers in general cases, this method may fail in the presence of strong specularity or noises. Besides, this method is sensitive to initial values and only guarantee finding a local minimum rather than a global minimum.~\cite{Related_23} assumes the specular highlight of a natural image has large intensity and applies a sparse and low-rank decomposition on the weighting matrices of specular and diffuse components. This approach is more robust to outliers than ~\cite{Related_22}. However, this method, similar to other methods in this category, reshapes the input color image into a $3 \times N$ matrix in which each column stores a pixel value. In that way, the spatial structure information of the original data will be lost, leading to inaccurate rank and low-efficiency computation.

In this paper, we address this issue by proposing a new tensor low-rank and sparse decomposition method, which keeps the spatial information. In our method, we also use the polarimetric cues to overcome the limitations of the chromaticity-based approaches, as discussed. 

\vspace{-7pt}\subsubsection{\textbf{Multi-Images Highlight Specular Separation Methods}}

Methods in the second category use multiple images to remove specular reflection from images. Since some highlights regions are direction-dependent, several methods use multiple images of one object (or scene) with different illumination directions~\cite{Related_9,Related_11,Related_12_1}, or from different point of views~\cite{Related_12_2,Related_12_3,Related_12_4,Related_13}, or with different polarization orientations~\cite{Related_15,Related_16,Related_17,Related_18}. Sato {\it et al.}~\cite{Related_11} employed the dichromatic model for separation by analyzing color in many images captured with a moving light source. Lin {\it et al.}~\cite{Related_12_1} also changed the light source direction to create two photometric images and used linear basis functions to separate the specular components. Multiple illumination based approaches are not applicable since the light source is usually fixed in the real world. 

Another early method uses images taken from different point of views to detect the specular regions based on the assumption of Lambertian consistency~\cite{Related_12_4}. Other methods use different viewing directions to remove the highlights by treating the specular pixels as outliers, and matching the remaining diffuse parts in other views~\cite{Related_12_2,Related_12_3}. However, These methods also fail if the size of the highlight region is large. In addition, methods in this category need a sequence of images which are not available in practice.

Different from the color-based methods, polarization-based methods use polarimetric cues for specular reflection separation. Nayar {\it et al.}~\cite{Related_18} proposed a method by analyzing the scene through the direct and global reflection components. Umeyama {\it et al.}~\cite{Related_16} used a fixed coefficient for specular components and separate them by applying independent component analysis (ICA). Wang {\it et al.}~\cite{Related_17} replaced the fixed coefficients with the spatially variable coefficient and improved the accuracy of the specular reflection separation. However, all these works still require a strict controllable light source, which limits the applicability of them in practice.

To address this issue, Wen {\it et al.}~\cite{Related_21} recently proposed a polarization guided specular reflection separation based on low-rank and sparse decomposition framework using a polarization camera. Since the polarization camera can capture all the polarization images in one shot, the method is more practical than all the above mentioned multi-image methods for real-world applications. Despite the promising results on some cases, they assume the diffuse components with different polarization angles remain constant, an assumption that is not the case in practice. They also followed the previous decomposition methods to cluster pixels with similar chromaticity values in a 2D matrix, where each row represents a pixel color (3 values). Therefore, the upper-bound of the rank for each cluster is 3, which may not be true for every scene. In addition, reshaping the input image into a $3 \times N$ matrix removes the spatial structural information of the original data and reduces the overall accuracy of the method. To address these issues, in this paper, we propose a polarimetric tensor low-rank and sparse decomposition framework which keeps the spatial structure of data, and uses them while recovering the diffuse image with the help of the phase angle regularization.

\subsection{Polarization Theory}
\label{pol_theory}
Our proposed method is based on a polarization camera, which implements pixel-level polarization filters and has resulted in real-time and high resolution measurment of incident polarization information. Each calculation unit of the camera consists of four pixels and uses four on-chip directional polarizers, at 0, 45, 90, and 135 degrees, to capture four perfectly aligned and polarized images. 

According to Fresnel's theory~\cite{Fresnel}, diffuse components $I_d$ remain roughly constant while the specular components $I_s$ vary under different polarization orientation. In the real world, the specular reflection is partially polarized and so part of the specular component $I_s$ also remains constant. Therefore, the intensity of a pixel across the different angles of polarization orientation $\vartheta$ is computed as follows~\cite{Related_21}.
\begin{equation}
\label{irradiance}
\begin{split}
I_{\phi}=I_d + I_{sc} + I_{sv}cos(2\vartheta-2\phi)
= I_c + I_{sv}cos(2\vartheta-2\phi)\hspace{25pt}
\end{split}
\end{equation}
where the specular component can be expressed as the sum of a constant component $I_{sc}$ and a cosine function term with amplitude $I_{sv}$.

Since four images are available from the polarization camera, Eq.~\ref{irradiance} forms an over-determined linear system of equations, and we are able to compute $\phi$, $I_c$, and $I_{sv}$. For an object with only diffuse reflection, phase angle $\phi$ determines the azimuth angle $\varphi$ up to a $\pi$ ambiguity~\cite{eccv16,shakeri2021}, as follows. 
\begin{equation}
\label{azimuth}
\varphi= \phi \,\,\, or \,\,\, \varphi = \phi + \pi
\end{equation}
where the azimuth angle $\varphi$ represents the angle between the projected surface normal direction and $x-$axis of the 2D image plane. We will use this property of $\phi$ to form the phase angle regularization term in Section~\ref{formulation}.

\section{proposed method}
\label{proposed_method}
Our proposed specular removal method has three main steps. In the first step, our method takes RGB polarization images as input. We then analyze them using the polarization theory to compute the polarization-based chromaticity image. In the second step, we create multiple representations for each image based on the obtained polarization chromaticity image and then form a tensor. Finally, we define a polarization regularization term and introduce a new tensor low-rank and sparse decomposition formulation using the polarization regularization. By solving the proposed formulation, we are able to extract the diffuse image from the input images.
 
Since the specular chromaticity could be assumed to be uniform for a given image and equal the chromaticity of the incident illumination~\cite{Related_24,Related_23}, we first estimate the illumination chromaticity $\Gamma$ of the captured image using~\cite{Related_25}, and then normalize the input image by $I(p)/(3\Gamma)$ such that $\Gamma_r = \Gamma_g = \Gamma_b = 1/3$ as a pure white illumination color. This preprocessing step is not necessary for computing the polarization-based chromaticity image; however, we use it to have a better initialization as will be explained in Section~\ref{initialize}. 

\subsection{Polarization-based Chromaticity Image}
\label{chromaticity_image}

Chromaticity image reveals color information of the image and most of the specular reflection separation methods use an estimated chromaticity image to remove the specular reflection. However, almost all of them have a strong assumption on illumination in order to compute chromaticity image, and they suffer from color distortion in practice. A recent approach~\cite{Related_21} for computing the chromaticity image based on polarization information showed promising results without the aforementioned assumption. Regardless of the illumination and the color of the specular components, the polarization based chromaticity method can reveal the intrinsic diffuse reflection of the image much better than the conventional methods. In our proposed method, we also use the similar idea to compute the chromaticity image $I_{chro}$ as follows.
\begin{equation}
\label{raw_SD}
\begin{split}
I_{chro} = \frac{I_{rawD}}{\sum_{\theta}{I_{rawD,\theta}+\overline{I_{min}}}} \hspace{25pt} \,\,\,\, \\
\overline{I_{min}} = \frac{\sum_p {min(I_r(p),I_g(p),I_b(p))}}{N}
\end{split}
\end{equation}
where $I_{rawS}=2I_{sv}$ and $I_{rawD}= I_c - I_{rawS}$ are the approximate specular and diffuse components, respectively. $\theta \in {R,G,B}$ and $\overline{I_{min}}$ is the average of the minimum values in the r, g, b channels of all pixels, and address the unstable situation caused by dark or noisy pixels~\cite{Related_21}.

\subsection{Multiple Representations and Initialization}
\label{initialize}
Unlike the existing methods which use a clustering algorithm to segment the input image, we propose to create a class of multiple candidates for each pixel with the help of the polarization-based chromaticity image. In particular, we use block-processing to select a certain number of candidates for each pixel. As explained in Section~\ref{chromaticity_image}, the chromaticity image can describe the scene without being affected by the specular reflection. Therefore, for a pixel $p$ at the center of each block, we randomly select $n_4$ pixels from that block with the similar intrinsic diffuse color so that $\|I_{chro}(p) - I_{chro}(q)\| < T$ is satisfied, where $q$ is a random pixel in the block and $T$ is a similarity threshold. Each of those $n_4$ selected pixels can be considered as one representation of pixel $p$. This process on all pixels provides $n_4$ different representation images for the observed image.

Now, we apply the chromaticity-based approach presented in~\cite{Related_8} to estimate the initial diffuse components for each representation. 
As discussed in Section~\ref{related_work}, current chromaticity-based approaches including~\cite{Related_8} may not be accurate in a large specular reflection region, and also do not provide satisfactory results on regions with one or more saturated color channels. However,~\cite{Related_8} still works in regions where the specular reflection is weak and so can provide a reasonable initial diffuse component for our optimization framework.


\subsection{Tensor Reconstruction and Formulation}
\label{formulation}

\begin{figure}[t]
\centering
  \includegraphics[width=\linewidth]{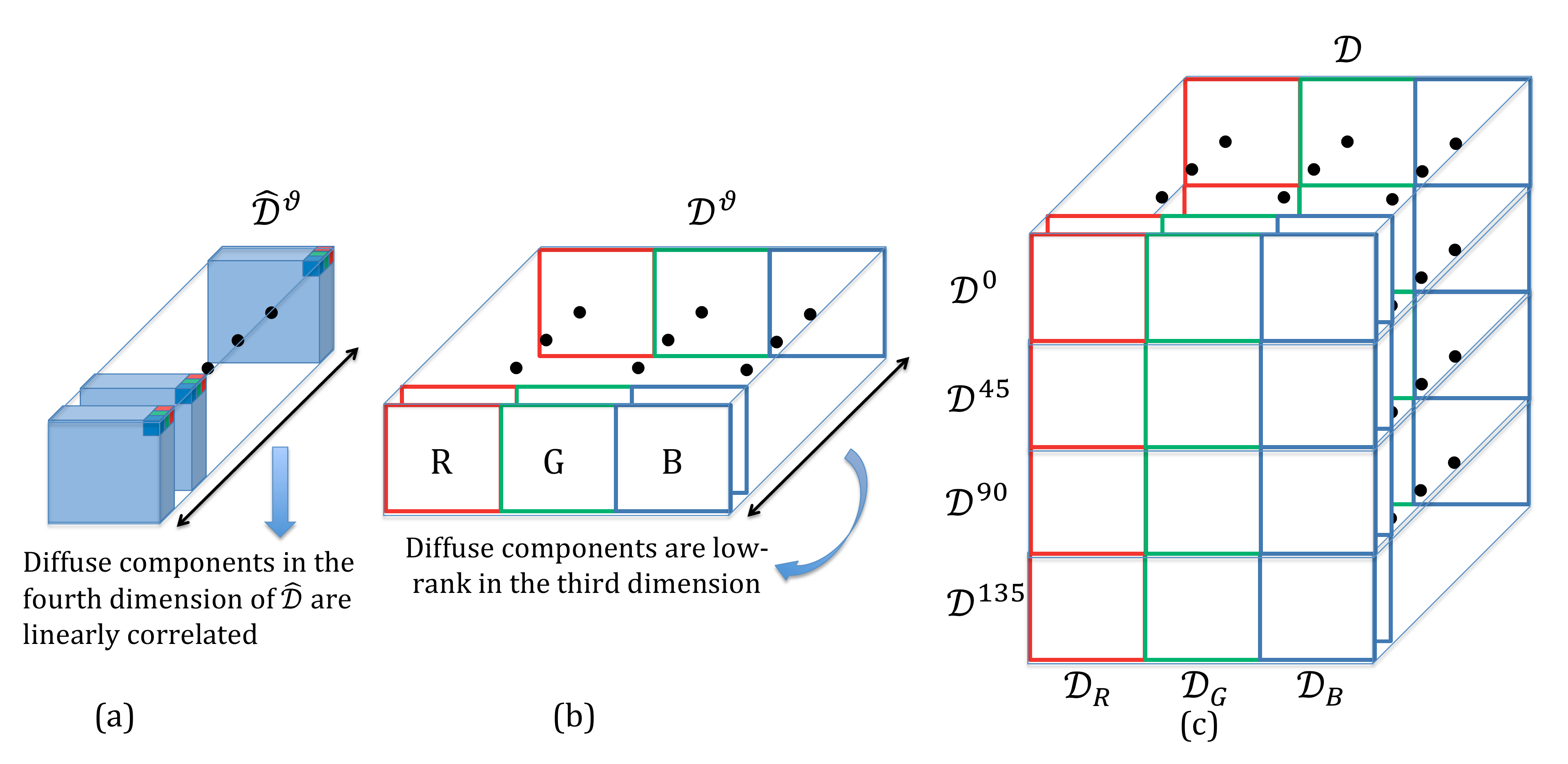}  
\caption{Our proposed tensor structures}
\label{fig:tensor_structure}
\end{figure}

Let $\mathcal{X}^{\vartheta} \in \mathbb{R}^{n_1 \times n_2 \times n_3 \times n_4}$ where $\mathcal{X}_i^{\vartheta} \in \mathbb{R}^{n_1 \times n_2 \times n_3}$ is the $i^{th}$ RGB representation of the original image ($i = 1, ..., n_4$) with the angle of polarization orientation $\vartheta \in \{0^{\circ}, 45^{\circ}, 90^{\circ}, 135^{\circ}\}$. $n_3 = 3$ denotes R, G and B channels. Therefore, $\hat{\mathcal{D}^{\vartheta}}=\Psi(\mathcal{X}^{\vartheta}) \in \mathbb{R}^{n_1 \times n_2 \times n_3 \times n_4}$ denotes the initial diffuse images obtained from Section~\ref{initialize} using~\cite{Related_8} (see Fig.~\ref{fig:tensor_structure}(a)). Corresponding pixels in the fourth dimension of the tensor $\mathcal{X}^{\vartheta}$ have a similar chromaticity basis color. This means the main structure of their diffuse component should be linearly correlated. Therefore the initial diffuse tensor $\mathcal{\hat{D}^{\vartheta}}$ should also be linearly correlated, and can be approximated by a small number of basis colors, corresponding to the rank in the tensor low-rank decomposition. Since $\hat{\mathcal{D}}^{\vartheta}$ is a $4D$-tensor, solving the decomposition is not straightforward. Although we can use low-rank tensor ring (TR) decomposition~\cite{Tensor_opt3} to separate the specular components from $\hat{\mathcal{D}}^{\vartheta}$, the method needs the TR rank to be specified before the optimization, which is not practical in real world objects and estimating it is time consuming. So, to make the problem tractable, we unfold $\hat{\mathcal{D}}^{\vartheta}$ along the $3^{rd}$ dimension as $\mathcal{D}^{\vartheta}= reshape(\hat{\mathcal{D}}^{\vartheta},[n_1,m, n_4])$ where $m=n_2 \times n_3$, so that $\mathcal{D}^{\vartheta} = [\mathcal{D_R}^{\vartheta},\mathcal{D_G}^{\vartheta},\mathcal{D_B}^{\vartheta}]$ (see Fig.~\ref{fig:tensor_structure}(b)). This block operation unfolds each of the color images by separating the channels and forms a 2D matrix. To take advantage of polarimetric cues in our tensor optimization, we stack all $\mathcal{D}^{\vartheta}, \vartheta = [0^{\circ}, 45^{\circ}, 90^{\circ}, 135^{\circ}]$ to form tensor $\mathcal{D}= [\mathcal{D_R,D_G,D_B}] \in \mathbb{R}^{k \times m \times n_4}, k= 4n_1$ which includes all the representations of all four polarization images (Fig.~\ref{fig:tensor_structure}(c)).


To separate specular reflection and saturated regions from diffuse components, we use the fact that specular components are spatially sparse in images. Based on this observation, regions with specular reflection should satisfy sparsity constraint, which can be modeled with the $l_{1}$-norm.

As discussed in Section~\ref{chromaticity_image}, using only chromaticity image and optimizing it to obtain diffuse image suffers from color distortion in practice. To address this issue and to improve the overall accuracy of the final diffuse image, we introduce a new regularization term based on polarimetric cues. This regularization term benefits from the similarity of phase angle between R, G, and B channels, and adds constraint between them which are already separated in $\mathcal{D}$. 

From Section~\ref{pol_theory}, $\phi$ is only related to the geometric surface information w.r.t the observer, and so it should remain unchanged among the color channels for each pixel in the ideal case. Based on this concept, $\phi^{\bar R}=\phi^{\bar G}=\phi^{\bar B}$ should be satisfied. We use $\phi^{\bar R} = mod(\phi^R, \pi)$ to avoid the $\pi$-ambiguity on azimuth angle. 

Now let $\Phi(\mathcal{D})$ compute $\phi^{\bar R}$, $\phi^{\bar G}$, and $\phi^{\bar B}$ from~(\ref{irradiance}) and updates $\phi=mean(\phi^{\bar R},$ $\phi^{\bar G},\phi^{\bar B})$. Also $\Phi^{-1}(\mathcal{D})$ is the inverse function of $\Phi(\mathcal{D})$ so that $\Phi^{-1}(\Phi({\mathcal{D}})) = \bar{\mathcal{D}}$ recovers the updated $\bar{\mathcal{D}}$ based on the new $\phi$.

With the above definition, we propose the following tensor low-rank and sparse decomposition formulation to separate the diffuse reflections from the specular reflection.
\begin{equation}
\label{formula}
\begin{split}
\min_{\mathcal{L},\mathcal{S}}\|\mathcal{L}\|_*+\lambda\|\tau\odot \mathcal{S}\|_{1}+ \gamma\|\Phi^{-1}(\Phi(\mathcal{L}))- \mathcal{L}\|_F^2 \\ 
s.t. \,\ \mathcal{D} = \mathcal{L}+\mathcal{S} \hspace{50pt}
\end{split}
\end{equation}
where $\odot$ denotes an element-wise multiplication. Since the tensor low-rank decomposition framework may smooth the images, we also define a spatially variant weight $\tau = \frac{1-I}{e^{-\alpha\|\nabla{I}\|^\beta}}$ where $I = \mathcal{D}(:,:,1)$, $\alpha=2$ and $\beta=0.25$ in our experiments. The third term in~(\ref{formula}) works as a phase angle regularization by updating $\mathcal{L}$ using~(\ref{irradiance}) and the above explanation so that the updated $\mathcal{L}$ shares a similar phase angle between color channels.

\subsection{Optimization}

In order to solve~(\ref{formula}), we use the inexact augmented Lagrangian method (IALM) with the augmented Lagrangian function $\mathcal{H}(\mathcal{L}, \mathcal{S},\mathcal{Y};\mu)$ whose main steps are described as follows.
\begin{equation}
\label{formula_relaxed}
\begin{split}
\mathcal{H}(\mathcal{L}, \mathcal{S},\mathcal{Y};\mu) = \|\mathcal{L}\|_*+\lambda\|\tau\odot \mathcal{S}\|_{1} + \gamma\|\Phi^{-1}(\Phi(\mathcal{L}))- \mathcal{L}\|_F^2\\
 + <\mathcal{Y}, \mathcal{D}-\mathcal{L}-\mathcal{S}>
+\frac{\mu}{2}\|\mathcal{D}-\mathcal{L}-\mathcal{S}\|_F^2 \hspace{20pt} 
\end{split}
\end{equation}
where $\mathcal{Y}$ is a Lagrangian multiplier, $\mu$ is a positive auto-adjusted scalar, and $<A,B>=trace(A^TB)$. $\lambda=1/\sqrt{max(n_1 \times n_2, n_3)n_4}$ and $\gamma$ is an increasing positive scalar. Now we first replace $\Phi^{-1}(\Phi(\mathcal{L}))$ with $\mathcal{Q}$, and then solve the problem through alternatively updating $\mathcal{L}$, $\mathcal{S}$, and $\mathcal{Q}$ in each iteration to minimize $\mathcal{H(L,S,Q,Y};\mu)$ with other variables fixed until convergence as follows.
\begin{equation}
\label{update_l}
\begin{split}
\mathcal{L}^{t+1}\leftarrow \min_{\mathcal{L}}\frac{1}{2\gamma+\mu}\|\mathcal{L}\|_*+\frac{1}{2}\|\mathcal{L}-(\frac{2\gamma}{2\gamma+\mu}(\mathcal{D}-\mathcal{S}^t)\\
+\frac{1}{2\gamma+\mu}\mathcal{Y}^t+\frac{\mu}{2\gamma+\mu}\mathcal{Q}^t)\|_F^2  
\end{split}
\end{equation}
\begin{equation}
\label{update_s}
\mathcal{S}^{t+1}\leftarrow \min_{\mathcal{S}}\lambda\|\tau\odot \mathcal{S}\|_{1}+\frac{\mu}{2}\|\mathcal{S}-(\mathcal{D}-\mathcal{L}^{t+1}+\frac{\mathcal{Y}^t}{\mu})\|_F^2
\end{equation}

\begin{equation}
\label{update_Q}
\mathcal{Q}^{t+1} = \Phi^{-1}(\Phi(\mathcal{L}^{t+1}))
\end{equation}
\begin{equation}
\label{update_Y}
\mathcal{Y}^{t+1} = \mathcal{Y}^t + \mu(\mathcal{D}^{t+1}-\mathcal{L}^{t+1}-\mathcal{S}^{t+1})
\end{equation}
where $\mu = min(\rho\mu,\mu_{max})$. Both~(\ref{update_l}) and~(\ref{update_s}) have closed form solutions in~\cite{Tensor_opt1} and~\cite{Tensor_opt2}, respectively. The error is computed as $\|\mathcal{D}-\mathcal{L}-\mathcal{S}\|_F /\|\mathcal{D}\|_F$ . The loop stops when the error falls below a threshold ($10^{-5}$ in our experiments). After convergence, $\mathcal{L}(:,:,1)$ includes all four diffuse polarization images. We use the average of those four results as the final diffuse image. 

\subsection{Time Complexity}
\label{time}
In this work, we use ADMM to update $\mathcal{L}$ and $\mathcal{S}$, which have closed form solutions. In these two steps the main cost lies in the update of $\mathcal{L}^{t+1} \in \mathbb{R}^{k \times m \times n_4}$, which requires computing FFT and $n_4$ SVDs of $k \times m$ matrices. Thus, time complexity of the first two steps per iteration is $O(kmn_4logn_4 +k_{(1)}k_{(2)}^2n_4)$, where $k_{(1)} = max(k, m)$ and $k_{(2)} = min(k, m)$~\cite{Tensor_opt1}. The time complexity to update $\mathcal{Q}^{t+1}$ is $O(kmn_4)$, since $\Phi(.)$ is an element-wise function. Therefore, the total time complexity of the optimization problem~(\ref{formula}) is $O(kmn_4logn_4 +k_{(1)}k_{(2)}^2n_4)$.


\begin{figure*}[t]
\centering
  \includegraphics[width=\linewidth]{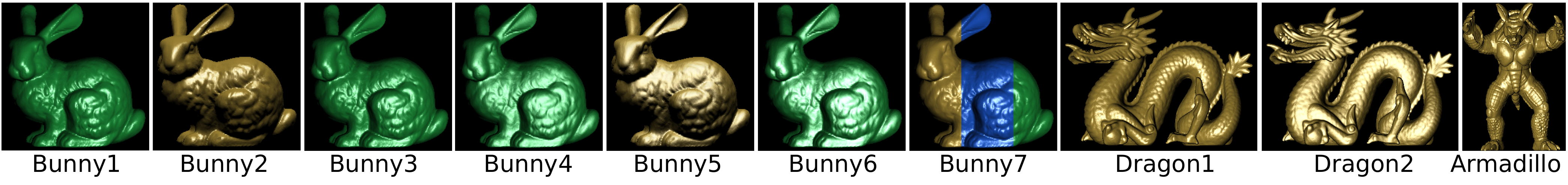}  
\caption{Sample synthetic images under different source of light and specularity information.}
\label{fig:sample_all}
\end{figure*}

\begin{table*}[t]
\caption{Quantitative evaluation in terms of PSNR and SSIM for Bunny images under different specular reflections (best scores: bold, second best scores: underline).}
\resizebox{\linewidth}{!}
{\begin{tabular}{|l|c|c|c|c|c|c|c|c|c|c|c|c|c|c|}
\hline
& \multicolumn{2}{|c|}{Bunny1} & \multicolumn{2}{|c|}{Bunny2} & \multicolumn{2}{|c|}{Bunny3} & \multicolumn{2}{|c|}{Bunny4} & \multicolumn{2}{|c|}{Bunny5} & \multicolumn{2}{|c|}{Bunny6} & \multicolumn{2}{|c|}{Bunny7}\\
\hline
& SSIM & PSNR & SSIM & PSNR & SSIM & PSNR & SSIM & PSNR & SSIM & PSNR & SSIM & PSNR & SSIM & PSNR\\
\hline
Yang'10~\cite{Related_8}     & 0.970 & 38.54 & 0.939 & 30.86 & \underline{0.961} & \underline{36.72} & 0.922 & 32.17 & 0.845 & 25.19 & 0.716 & 17.66 & 0.932 & \underline{31.02}\\
Shen'8~\cite{Related_8_1}      & 0.931 & 34.10 & 0.948 & \underline{32.35} & 0.857 & 28.86 & 0.616 & 19.25 & 0.804 & 23.77 & 0.492 & 16.94 & 0.874 & 28.03\\
Shen'9~\cite{Related_8_2}      & 0.932 & 35.88 & \underline{0.950} & 31.17 & 0.904 & 32.14 & 0.694 & 22.57 & 0.797 & 24.63 & 0.602 & 19.27 & 0.889 & 28.94\\
Shen'13~\cite{Related_8_4}     & \underline{0.971} & \underline{38.66} & 0.948 & 32.25 & 0.907 & 31.51 & 0.569 & 16.24 & 0.701 & 18.89 & 0.483 & 15.52 & 0.847 & 25.99\\
Tan'05~\cite{Related_4}      & 0.905 & 30.46 & 0.865 & 26.02 & 0.888 & 29.59 & 0.847 & 27.47 & 0.769 & 21.82 & \underline{0.717} & \underline{20.90} & 0.752 & 22.12\\
Yoon'06~\cite{Related_8_3}     & 0.897 & 25.30 & 0.949 & 32.28 & 0.874 & 23.35 & 0.777 & 19.75 & \underline{0.871} & \underline{26.01} & 0.658 & 13.83 & 0.861 & 22.59\\
Akashi'16~\cite{Related_22}   & 0.766 & 21.70 & 0.866 & 24.44 & 0.701 & 19.28 & 0.551 & 14.06 & 0.686 & 16.11 & 0.491 & 12.08 & 0.850 & 24.05\\
Yamamoto'19~\cite{Related_8_5} & 0.969 & 38.51 & 0.938 & 30.83 & 0.960 & 36.71 & \underline{0.923} & \underline{32.18} & 0.844 & 25.19 & 0.716 & 17.66 & \underline{0.933} & 31.01\\
Gang'21~\cite{Related_22_1}     & 0.854 & 27.20 & 0.852 & 12.76 & 0.855 & 25.95 & 0.853 & 22.07 & 0.691 & 18.25 & 0.841 & 19.09 & 0.849 & 29.08\\
Ours        & \textbf{0.993} & \textbf{48.11} & \textbf{0.981} & \textbf{41.77} & \textbf{0.989} & \textbf{45.72} & \textbf{0.966} & \textbf{36.67} & \textbf{0.951} & \textbf{33.90} & \textbf{0.954} & \textbf{35.16} & \textbf{0.978} & \textbf{38.26}\\
\hline
\end{tabular}}
\label{quantitative}
\end{table*}

\section{Experimental Results and Discussion}
\label{experiment}
In this section we present the experimental results of our proposed method on both synthetic and real polarization images. In the first set of experiments, we evaluate our method quantitatively by comparing the results with those from chromaticity-based approaches~\cite{Related_8,Related_8_1,Related_8_2,Related_8_3,Related_8_4,Related_8_5,Related_4}, a matrix factorization based mothod~\cite{Related_22}, and a deep learning based method~\cite{Related_22_1} on synthetic polarization images. In the second set of experiments we show the qualitative results of the proposed method and compare them with the results of the competing methods on real polarization images captured by a polarization camera.

\subsection{Evaluation on Synthetic Polarization Images}
To overcome the lack of polarization image datasets for highlight specular reflection removal, we have had to first create our own synthetic polarization image datasets, and then evaluate our proposed method on them. To create our synthetic dataset, we use point clouds of three objects ``Bunny", ``Dragon", and ``Armadillo" from the Stanford 3D scanning repository~\cite{stanford}, and create 2D depth images. We then render the depth images with the Blinn-Phong reflectance model under different arbitrary point lighting source $s$ using the pinhole camera model, with uniform and non-uniform albedo texture. In this process we use polarizer filter angles at $0^{\circ}, 45^{\circ}, 90^{\circ},$ and $135^{\circ}$ to create synthetic polarization images. Fig.~\ref{fig:sample_all} shows the obtained images under different sources of light and specularity information which we use in our evaluations. Each image in Fig.~\ref{fig:sample_all} is the average of the four polarization images.  

\begin{figure}[b]
\centering
  \includegraphics[width=\linewidth]{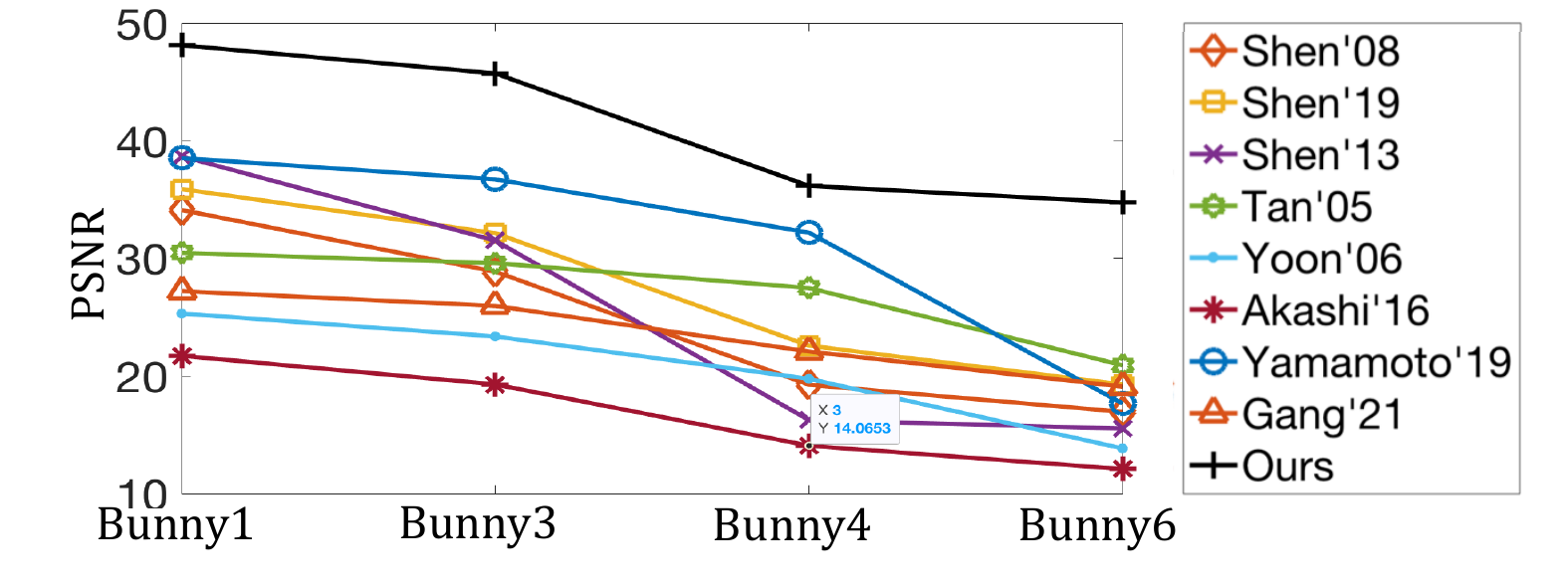}  
  \vspace{-10pt}
\caption{Performance of the proposed method and the competing methods while the specular reflection increases}
\label{fig:saturate_eval}
\end{figure}

In the first experiment, we evaluate the results of our proposed method and compare the results with the competing methods using two common evaluation metrics SSIM~\cite{SSIM} and PSNR. Table~\ref{quantitative} shows the performance comparison on ``Bunny" dataset under different illumination and specular reflections. For all the samples, the proposed method achieves the best scores with significant improvement in the evaluation metrics. In images with weak specular reflection, almost all competing methods provide acceptable results; however, the proposed method still outperforms all of them. The capability of the proposed method is more clear in cases with strong specular reflection, e.g, ``Bunny5" and ``Bunny6", where the competing methods cannot recover the diffuse reflection properly. Experiment on ``Bunny7" evaluates our proposed method in the presence of saturated regions under varying albedo which shows a noticeable improvement in comparison with the competitive methods.

We also evaluate the proposed method while the specular reflection increases in Fig.~\ref{fig:saturate_eval}. Specular reflection and saturation regions are increasing in the samples in the x-axis. This figure specifically visualizes the capability of our proposed method in the presence of strong specular reflection and saturation (e.g., ``Bunny6"). 

\begin{figure*}[t]
\centering
  \includegraphics[width=\linewidth]{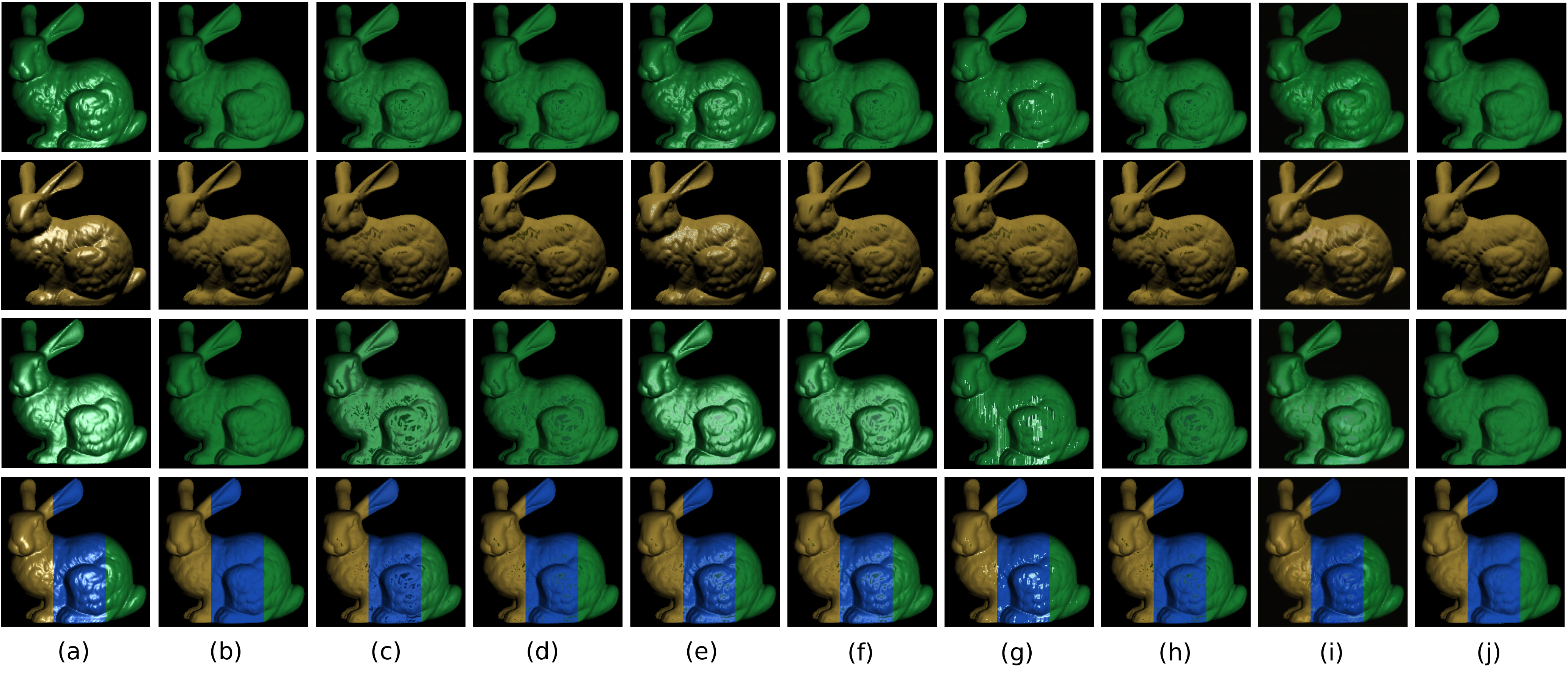}  
\caption{Comparison of results between the proposed method and the competing methods. (a) RGB image, (b) Ground truth, (c) Shen'08~\cite{Related_8_1}, (d) Yang\cite{Related_8}, (e) Akashi~\cite{Related_22}, (f) Shen'13~\cite{Related_8_4}, (g) Yoon~\cite{Related_8_3}, (h) Yamamoto~\cite{Related_8_5}, (i) Gang~\cite{Related_22_1}, (j) Ours}
\label{fig:bunny_all}
\vspace{-10pt}
\end{figure*}

Fig.~\ref{fig:bunny_all} shows the qualitative results of four sample images from Table~\ref{quantitative}. This figure clearly shows that the chromaticity based methods may not have enough information to recover the diffuse image in regions with strong specular reflection, and all of them fail in such cases. In contrast, the proposed method can recover diffuse values of those regions due to use of polarimetric cues and spatial information in the proposed tensor structure. 

\begin{figure}[b]
\centering
  \includegraphics[width=\linewidth]{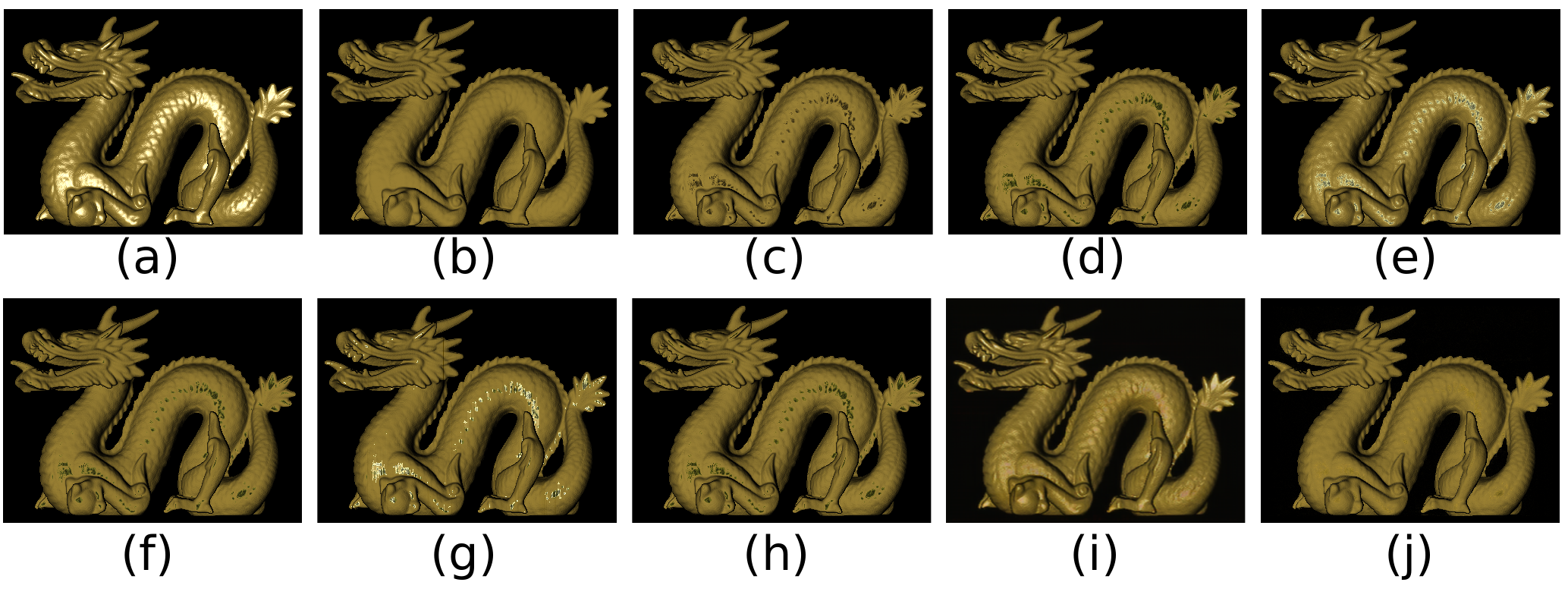}  
\caption{Comparison of results between the proposed method and the competing methods. (a) RGB image, (b) Ground truth, (c) Shen'08~\cite{Related_8_1}, (d) Yang\cite{Related_8}, (e) Akashi~\cite{Related_22}, (f) Shen'13~\cite{Related_8_4}, (g) Yoon~\cite{Related_8_3}, (h) Yamamoto~\cite{Related_8_5}, (i) Gang~\cite{Related_22_1}, (j) Ours}
\label{fig:dragon}
\end{figure}

\begin{figure}[b]
\centering
  \includegraphics[width=\linewidth]{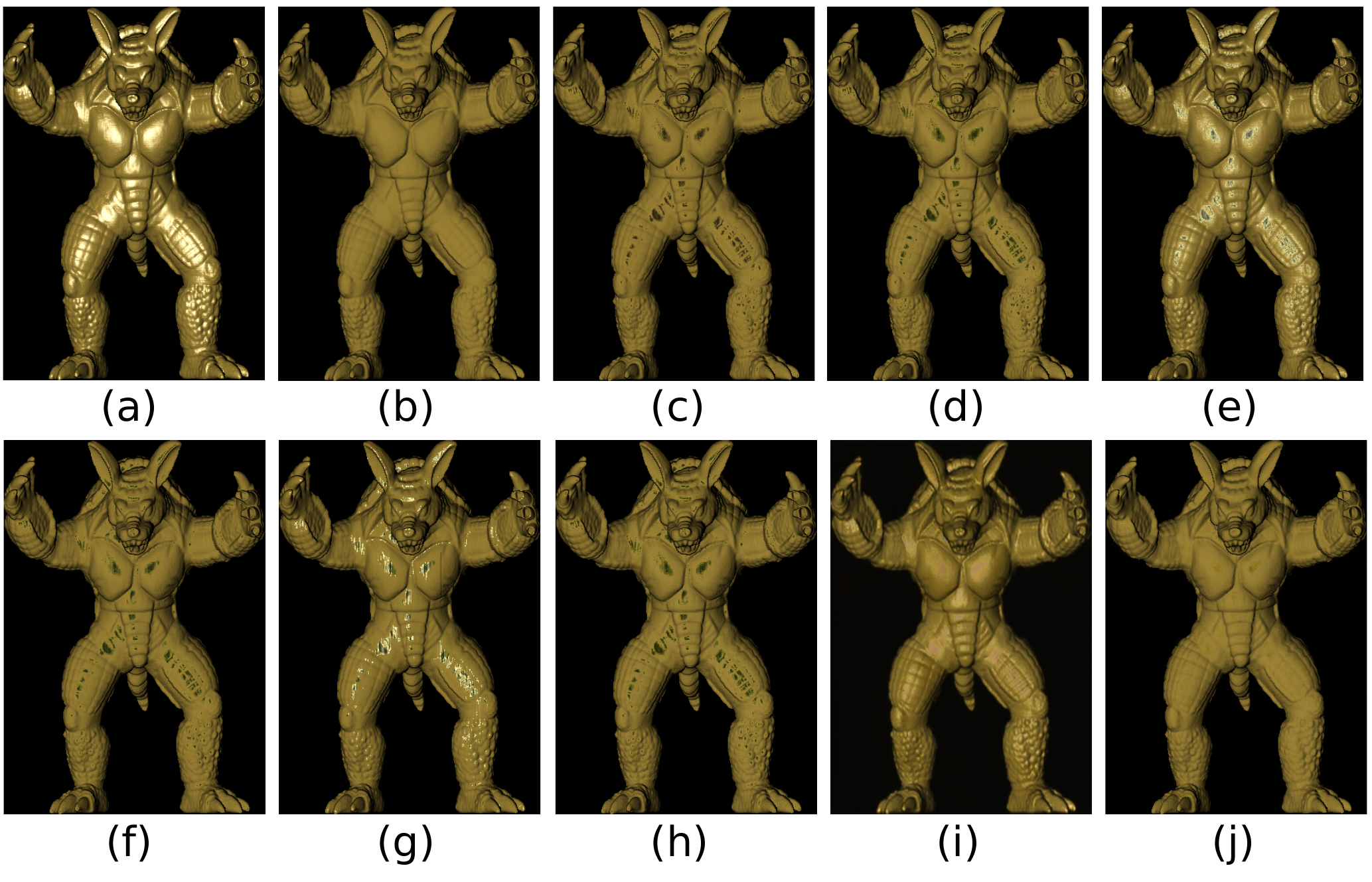}  
\vspace{-10pt}
\caption{Comparison of results between the proposed method and the competing methods. (a) RGB image, (b) Ground truth, (c) Shen'08~\cite{Related_8_1}, (d) Yang\cite{Related_8}, (e) Akashi~\cite{Related_22}, (f) Shen'13~\cite{Related_8_4}, (g) Yoon~\cite{Related_8_3}, (h) Yamamoto~\cite{Related_8_5}, (i) Gang~\cite{Related_22_1}, (j) Ours}
\label{fig:armadillo}
\end{figure}

To show the capability of our proposed method, we also evaluate our method with two more datasets ``Dragon" and ``Armadillo" under different conditions with saturated regions, quantitatively. Table~\ref{dragon_armadillo} shows the performance of the proposed method in comparison with the competing methods. In these experiments, the proposed method also outperforms all the competing methods in terms of both SSIM and PSNR. Figs.~\ref{fig:dragon} and~\ref{fig:armadillo} show the qualitative results of all the methods available in Table~\ref{dragon_armadillo} on ``Dragon1" and ``Armadillo". 

\begin{table}[b]
\centering
\caption{Quantitative evaluation in terms of PSNR and SSIM for Bunny images under different specular reflections (best scores: bold, second best scores: underline).}
\resizebox{\linewidth}{!}
{\begin{tabular}{|l|c|c|c|c|c|c|}
\hline
& \multicolumn{2}{|c|}{Dragon1} & \multicolumn{2}{|c|}{Dragan2} & \multicolumn{2}{|c|}{Armadillo} \\
\hline
& SSIM & PSNR & SSIM & PSNR & SSIM & PSNR\\
\hline
Yang'10~\cite{Related_8}     & 0.910 & 28.32 & 0.727 & 21.58 & 0.886 & 27.64\\
Shen'8~\cite{Related_8_1}      & 0.921 & 29.12 & 0.662 & 19.01 & 0.905 & 28.41\\
Shen'9~\cite{Related_8_2}      & \underline{0.941} & \underline{30.95} & 0.721 & \underline{22.68} & \underline{0.929} & \underline{30.22}\\
Shen'13~\cite{Related_8_4}     & 0.924 & 29.12 & 0.610 & 17.47 & 0.904 & 28.34\\
Tan'05~\cite{Related_4}      & 0.838 & 24.64 & 0.682 & 20.06 & 0.814 & 23.66\\
Yoon'06~\cite{Related_8_3}     & 0.884 & 24.81 & 0.745 & 20.32 & 0.851 & 24.02\\
Akashi'16~\cite{Related_22}   & 0.856 & 25.51 & 0.603 & 15.08 & 0.804 & 22.92\\
Yamamoto'19~\cite{Related_8_5} & 0.910 & 28.32 & 0.727 & 21.59 & 0.886 & 27.64\\
Gang'21~\cite{Related_22_1}     & 0.756 & 26.27 & \underline{0.753} & 19.20 & 0.788 & 25.75\\
Ours        & \textbf{0.975} & \textbf{38.70} & \textbf{0.906} & \textbf{30.21} & \textbf{0.971} & \textbf{38.88}\\
\hline
\end{tabular}}
\label{dragon_armadillo}
\end{table}

\subsection{Evaluation of the phase angle regularization and $\tau$}
In this section, we evaluate the effects of the penalty term $\tau$ and the phase angle regularization term on our proposed tensor structure. In the first experiment, we evaluate the capability of our proposed tensor decomposition without phase angle regularization terms, where~(\ref{formula}) becomes
\begin{equation}
\label{formula_2}
\begin{split}
\min_{\mathcal{L},\mathcal{S}}\|\mathcal{L}\|_*+\lambda\|\tau\odot\mathcal{S}\|_{1} \,\,\,\,
s.t. \,\ \mathcal{D} = \mathcal{L}+\mathcal{S} 
\end{split}
\end{equation}
We further remove the penalty term $\tau$ to show the capability of the proposed tensor structure itself, where the proposed formulation becomes
\begin{equation}
\label{formula_1}
\begin{split}
\min_{\mathcal{L},\mathcal{S}}\|\mathcal{L}\|_*+\lambda\|\mathcal{S}\|_{1} \,\,\,\,
s.t. \,\ \mathcal{D} = \mathcal{L}+\mathcal{S} 
\end{split}
\end{equation}
We solve both~(\ref{formula_2}) and~(\ref{formula_1}) and compare the obtained results with the results of~(\ref{formula}). Table~\ref{ablation} shows that~(\ref{formula_1}) without $\tau$ and polarization regularization term can still outperform the competing methods of Table~\ref{quantitative}, due to the use of tensor structure which keeps the spatial information of the images. The results of~(\ref{formula_2}) in Table~\ref{ablation} show the effectiveness of the spatially variant weight $\tau$ where prevents the smoothing of the images through the optimization process, a common artifact of low-rank decomposition. $\tau$ works as a penalty term and increases the cost of optimization around high frequency component of images (e.g., edges) and therefore, those regions grouped into the low-rank component $\mathcal{L}$. This means, $\mathcal{L}$ includes all the details of the images as much as possible which can boost the accuracy of the recovered diffuse images. Finally, the last two columns of Table~\ref{ablation} show the results of our proposed formulation presented in~(\ref{formula}). This experiment illustrates the benefit of polarimetric information to improve the accuracy of specular reflection separation from diffuse images, especially in regions of strong specular reflection or in saturated areas (e.g., ``Bunny6"). 

\begin{figure*}
\centering
  \includegraphics[width=\linewidth]{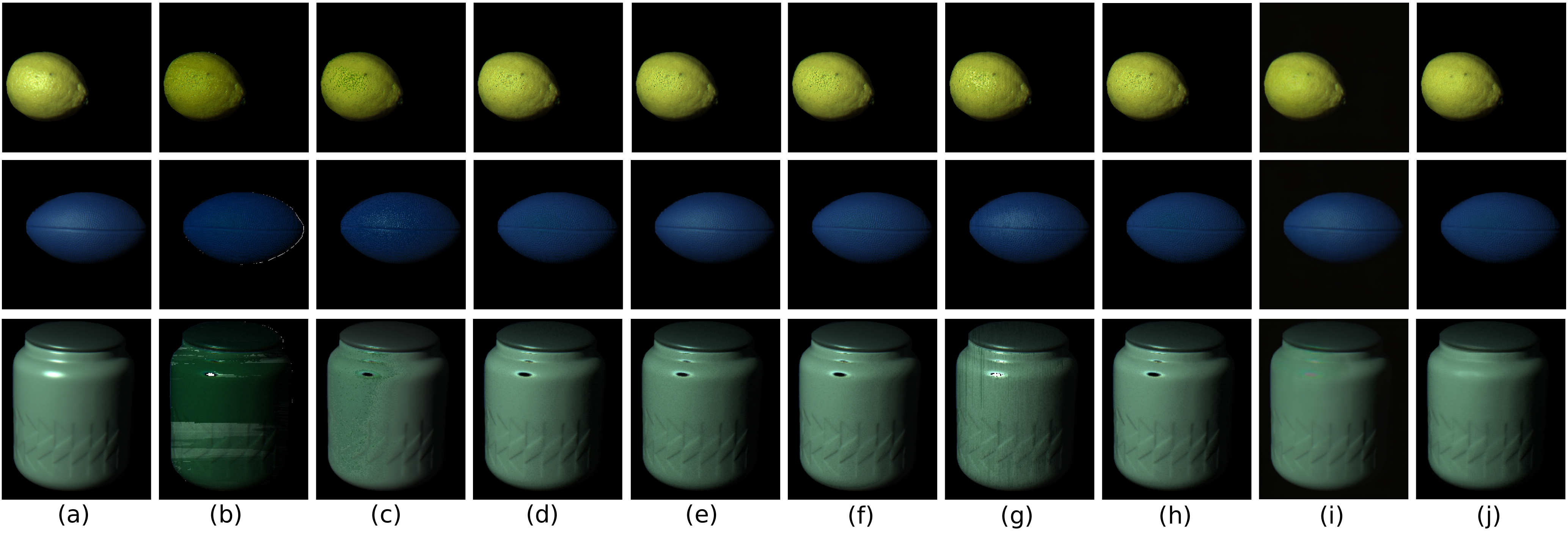}  
\caption{Comparison of results between the proposed method and the competing methods. (a) RGB image, (b) Tan~\cite{Related_4}, (c) Shen'08~\cite{Related_8_1}, (d) Yang\cite{Related_8}, (e) Akashi~\cite{Related_22}, (f) Shen'13~\cite{Related_8_4}, (g) Yoon~\cite{Related_8_3}, (h) Yamamoto~\cite{Related_8_5}, (i) Gang~\cite{Related_22_1}, (j) Ours}
\label{fig:real}
\end{figure*} 

\subsection{Evaluation on Real Polarization Images}
In this experiment, we show the results of the proposed method on real polarization images captured by a polarization camera. Since there is no available ground truth for these polarization images, we only compare our method with the competing methods qualitatively in Fig.~\ref{fig:real}. Overall, our method performs well on a diversity of real images which may contain various materials with different textures, overexposure and natural illumination. As shown, our method can handle saturation areas without annoying artifacts. In comparison with the competetive methods, the proposed method well suppresses the specular highlights and preserves the image details as much as possible. 

\begin{table}[t]
\centering
\caption{Evaluating the effects of the penalty term $\tau$ and the phase angle regularization in terms of PSNR and SSIM for ``Bunny4" and ``Bunny6" images with strong specular reflections}
\resizebox{\linewidth}{!}
{\begin{tabular}{|c|c|c|c|c|c|c|}
\hline
& \multicolumn{2}{|c|}{Ours w/o phase angle term} & \multicolumn{2}{|c|}{Ours w/o phase angle term} & \multicolumn{2}{|c|}{Ours} \\
& \multicolumn{2}{|c|}{and $\tau$ (Results of~(\ref{formula_1}))} & \multicolumn{2}{|c|}{Results of~(\ref{formula_2})} & \multicolumn{2}{|c|}{Results of~(\ref{formula})} \\
\hline
& \hspace{12pt}SSIM\hspace{12pt} & PSNR & \hspace{12pt}SSIM\hspace{12pt} & PSNR & \hspace{3pt}SSIM\hspace{3pt} & PSNR\\
\hline
Bunny4     & 0.921 & 31.45 & 0.944 & 34.87 & 0.966 & 36.67\\
Bunny6      & 0.8141 & 23.36 & 0.902 & 31.01 & 0.954 & 35.16\\
\hline
\end{tabular}}
\label{ablation}
\end{table}

\section{Conclusion}
\label{conclusion}
In this paper we have proposed a novel method based on polarimetric information for highlight specular reflection removal. Our method exploits polarimetric cues obtained by a polarization camera to extract the diffuse image from the polarization images. The proposed method is built upon a tensor low-rank and sparse decomposition framework. In our method, we first select multiple candidates for each pixel based on the polarization chromaticity image. Then we stack those candidates in a tensor structure where the candidates for each pixel can be considered as a different representation of that pixel. Since the diffuse colors of pixels in a small area with similar material and chromaticity value are linearly correlated, our proposed tensor low-rank method can recover them. Different from previous low-rank decomposition based methods, the proposed method keeps the spatial structure of data and recovers the saturated regions using diffuse color of adjacent pixels. Due to the use of phase angle regularization between the color channels, the proposed method also can handle color distortion without annoying artifacts that are common with chromaticity-based approaches. Our experimental results on both synthetic and real polarization images demonstrate the superiority of our method with respect to the state-of-the-art.

{\small
\bibliographystyle{ieee_fullname}
\bibliography{egbib}
}

\end{document}